\documentclass{article} 
\usepackage{colm2024_conference}
\usepackage{microtype}
\usepackage{hyperref}
\usepackage{url}
\usepackage{booktabs}
\usepackage{CJKutf8}
\usepackage{graphicx}
\usepackage{multicol}
\usepackage{multirow}
\usepackage{array}
\usepackage{amsmath}
\usepackage{booktabs}
\usepackage{enumitem}
\usepackage{wrapfig}
\usepackage{algorithm}
\usepackage{algpseudocode}
\usepackage{microtype}
\usepackage{amsmath}
\usepackage{colortbl}
\usepackage[utf8]{inputenc}
\definecolor{lightgray}{rgb}{0.9,0.9,0.9}
\usepackage{caption}
\usepackage{subcaption}
\usepackage{xcolor}
\usepackage{setspace}
\usepackage{url}
\usepackage{colortbl}
\usepackage{tabularx}
\usepackage{blindtext}
\usepackage{pgfplots}
\pgfplotsset{compat=1.18} 
\usepackage{tikz}
\usetikzlibrary{er,positioning,bayesnet}
\usepackage{makecell}
\usepackage{tipa}
\usepackage{siunitx}
\usepackage{nicefrac}
\usepackage{tocloft}
\usepackage{listings}
\usepackage{tcolorbox}
\usepackage{xltabular}
\usepackage{adjustbox}
\usepackage{xurl}
\usepackage{setspace}
\usepackage{lipsum}

\title{HY-MT1.5 Technical Report}

\author{
	\bf \large Tencent Hunyuan Team
}
\begin{document}
\begin{CJK*}{UTF8}{gbsn}
\maketitle
		
\begin{abstract}
In this report, we introduce our latest translation models, \textbf{HY-MT1.5-1.8B} and \textbf{HY-MT1.5-7B}, a new family of machine translation models developed through a holistic training framework tailored for high-performance translation. Our methodology orchestrates a multi-stage pipeline that integrates general and MT-oriented pre-training, supervised fine-tuning, on-policy distillation, and reinforcement learning. \textbf{HY-MT1.5-1.8B}, the 1.8B-parameter model demonstrates remarkable parameter efficiency, comprehensively outperforming significantly larger open-source baselines (e.g., Tower-Plus-72B, Qwen3-32B) and mainstream commercial APIs (e.g., Microsoft Translator, Doubao Translator) in standard Chinese-foreign and English-foreign tasks. It achieves approximately 90\% of the performance of ultra-large proprietary models such as Gemini-3.0-Pro, while marginally trailing Gemini-3.0-Pro on WMT25 and Mandarin-minority language benchmarks, it maintains a substantial lead over other competing models. Furthermore, \textbf{HY-MT1.5-7B} establishes a new state-of-the-art for its size class, achieving 95\% of Gemini-3.0-Pro’s performance on Flores-200 and surpassing it on the challenging WMT25 and Mandarin-minority language test sets. Beyond standard translation, the HY-MT1.5 series supports advanced constraints, including terminology intervention, context-aware translation, and format preservation. Extensive empirical evaluations confirm that both models offer highly competitive, robust solutions for general and specialized translation tasks within their respective parameter scales.



\small {\color{blue}\textbf{HY-MT1.5-1.8B}}: \url{https://huggingface.co/tencent/HY-MT1.5-1.8B}

\small {\color{blue}\textbf{HY-MT1.5-7B}}: \url{https://huggingface.co/tencent/HY-MT1.5-7B}

\normalsize {\color{blue}\textbf{Code Repository}}: \url{https://github.com/Tencent-Hunyuan/HY-MT}

\end{abstract}
	
\begin{figure}[h]
\centering
\includegraphics[width=1\linewidth]{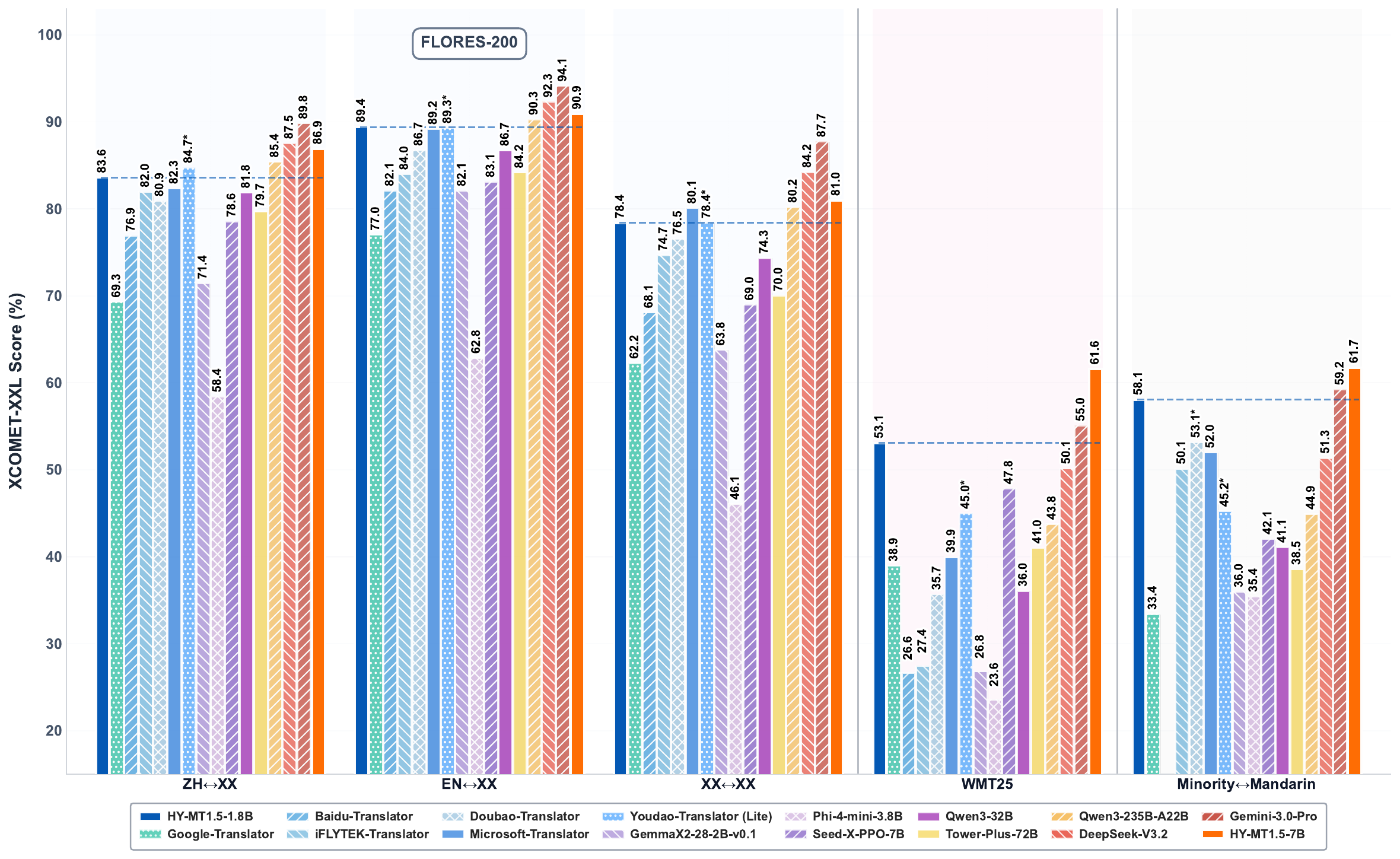}
\caption{Benchmark performance of HY-MT1.5 models and state-of-the-art baselines.}
\label{fig:overall}
\end{figure}		

\begin{figure}[h]
    \centering 
    \begin{subfigure}{0.48\textwidth}
        \centering 
        \includegraphics[width=\linewidth]{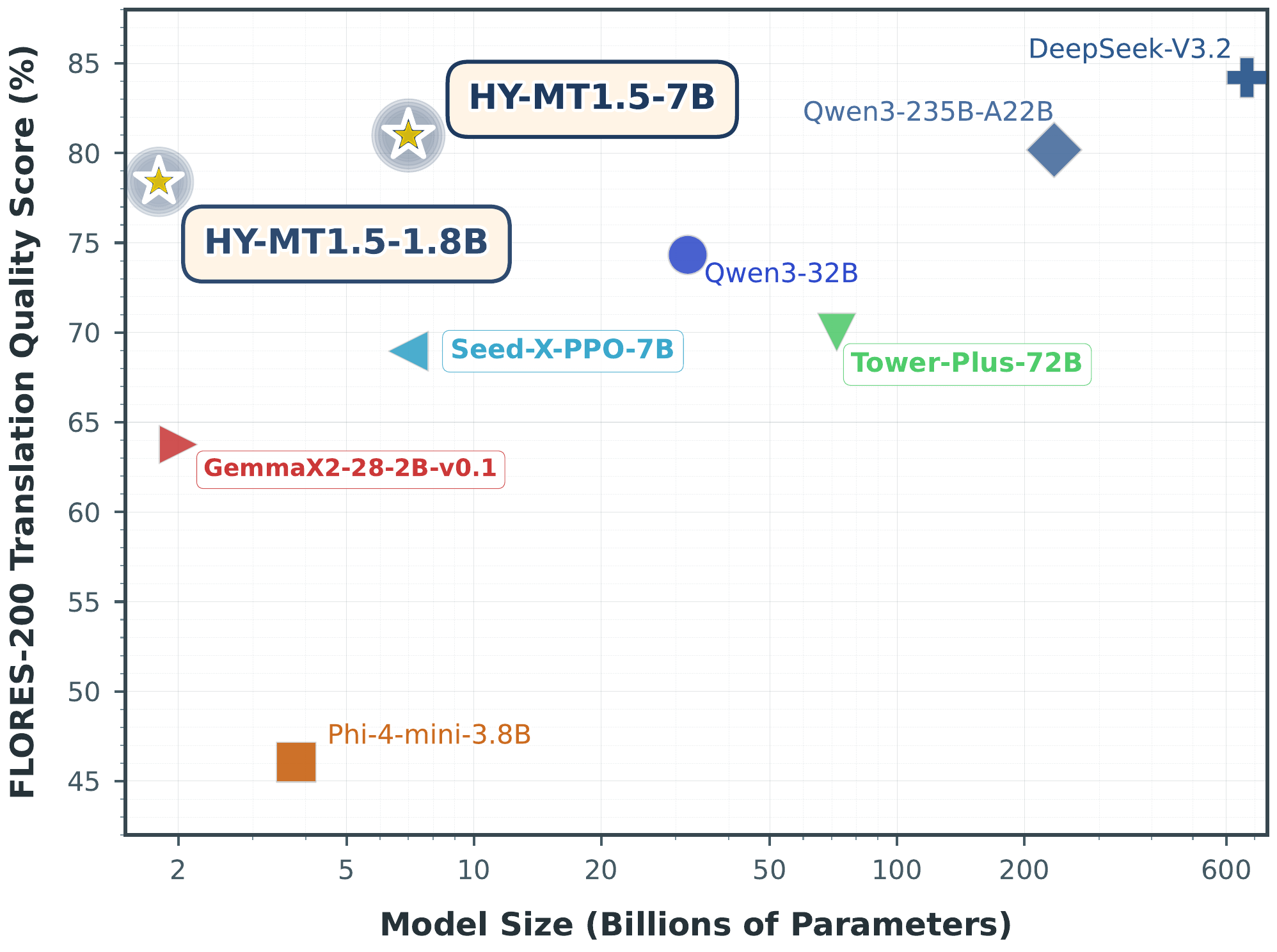} 
        \subcaption{Model size versus Flores-200 (XX \( \Leftrightarrow \) XX) translation quality for different-scale open-source models.} 
        \label{subfig:flores200} 
    \end{subfigure}
    \hfill 
    \begin{subfigure}{0.48\textwidth}
        \centering
        \includegraphics[width=\linewidth]{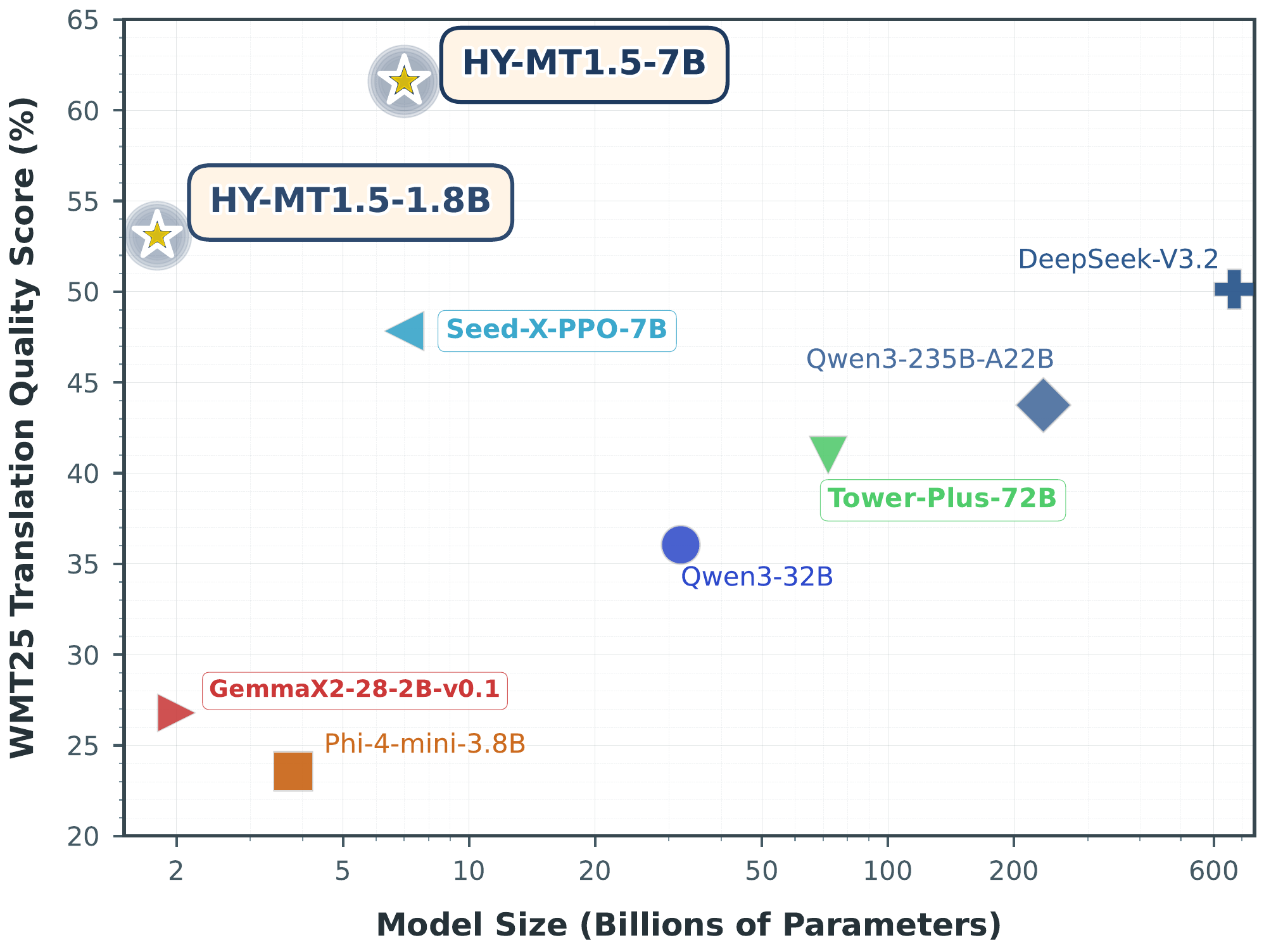}
        \subcaption{Model size versus WMT25 translation quality for different-scale open-source models.}
        \label{subfig:wmt25}
    \end{subfigure}
    \caption{Comparison of model size versus translation quality across Flores-200 and WMT25 datasets for open-source models.} 
    \label{fig:total} 
\end{figure}

\section{Introduction}

Machine translation (MT) has long been a high-demand practical goal and a prominent research challenge pursued by the computational linguistics community over the past few decades \citep{brown1990statistical, brown1993mathematics, bleu, sutskever2014, bahdanau2015, wu2016google, vaswani2017attention}. The rapid advancement of large language models (LLMs) has revolutionized the learning paradigm of machine translation, shifting from traditional rule-based and statistical approaches to large-scale neural learning methodologies, and continuously pushing the boundaries of translation quality to unprecedented levels \citep{zhu2024,kocmi2024,pang2025}. State-of-the-art closed-source models, such as Gemini-3.0-Pro \citep{gemini_3pro}, have demonstrated capabilities that approach or surpass those of expert human translators in specific language pairs.

Nevertheless, significant challenges persist in machine translation \citep{kocmi-etal-2025-findings}. First and foremost, the balance between translation quality and efficiency remains a critical unaddressed issue. State-of-the-art large-scale closed-source models often deliver high translation quality but incur prohibitive deployment costs and low inference efficiency due to their enormous parameter sizes, making them inaccessible for widespread practical applications (e.g., edge device deployment and high-throughput translation scenarios). Meanwhile, existing lightweight open-source models typically sacrifice translation quality to achieve efficiency, failing to match the performance of large closed-source models, thereby exacerbating the disparity between practical availability and the demands for high-quality translation. 
Second, current translation systems are predominantly limited to basic text translation tasks and lack support for customized translation requirements through flexible interaction with prompts. For instance, key capabilities such as contextual translation (maintaining coherence across multi-turn or long-document translation) and formatted translation (preserving the original document structure, such as tables, lists, and formulas) are insufficiently addressed. These limitations hinder the adaptation of translation systems to diverse real-world scenarios, in which customized requirements are increasingly prevalent.

These two core challenges—quality-efficiency imbalance and inadequate customized translation support—severely restrict the further advancement and widespread adoption of machine translation technology, highlighting the urgent need for innovative solutions that can simultaneously address efficiency, quality, and customization demands.

To directly tackle the aforementioned two core challenges, we present the HY-MT1.5 models and corresponding technical solutions, with three key contributions that closely align with the pain points:
\begin{enumerate}[left=2pt, itemsep=2pt, topsep=4pt]

\item \textbf{High-performance and efficient HY-MT1.5 models}: Targeting the core challenge of balancing translation quality and efficiency, we propose HY-MT1.5-1.8B and HY-MT1.5-7B that achieve a superior performance-efficiency balance. As shown in Figure \ref{fig:overall} and \ref{fig:total}, the 1.8B-parameter HY-MT1.5-1.8B comprehensively outperforms mainstream medium-sized open-source models (e.g., Tower-Plus-72B \citealp{towerplus}, Qwen3-32B \citep{qwen3}) and commercial translation APIs, reaching the 90th percentile of ultra-large closed-source models like Gemini-3.0-Pro \citep{gemini_3pro}. The 7B-parameter HY-MT1.5-7B further reaches the 95th percentile of Gemini-3.0-Pro \citep{gemini_3pro} on the Flores-200 dataset \citep{flores-200} and even surpasses it on WMT25 \citep{kocmi-etal-2025-findings} and Mandarin-minority language test sets, while maintaining efficient deployment capabilities that facilitate widespread practical application.

\item \textbf{Holistic and effective training scheme}: We develop a tailored training framework for machine translation, integrating general pre-training, MT-oriented pre-training, supervised finetuning, on-policy distillation, and reinforcement learning. This framework enables the models to excel in both general and low-resource translation scenarios, laying a solid foundation for the superior performance of the HY-MT1.5 models.

\item \textbf{Practical distinctive features}: Addressing the limitation of insufficient customized translation support in existing systems, the HY-MT1.5 models are equipped with practical features including terminology intervention, contextual translation, and formatted translation. These capabilities enable flexible response to customized translation demands through prompt interactions, significantly enhancing adaptability to diverse real-world application scenarios and bridging the gap between academic performance and industrial needs.
\end{enumerate}


The remainder of this report is organized as follows: we first elaborate on the holistic training framework and development details of the HY-MT1.5 models. Subsequently, we present extensive experimental evaluations to validate the models' performance across various representative translation benchmarks, focusing on their performance-efficiency balance and customized translation capabilities. Finally, we discuss key findings and outline future research directions.

\section{Methodology}

\begin{figure}[ht!]
	\centering
	\includegraphics[width=1\linewidth]{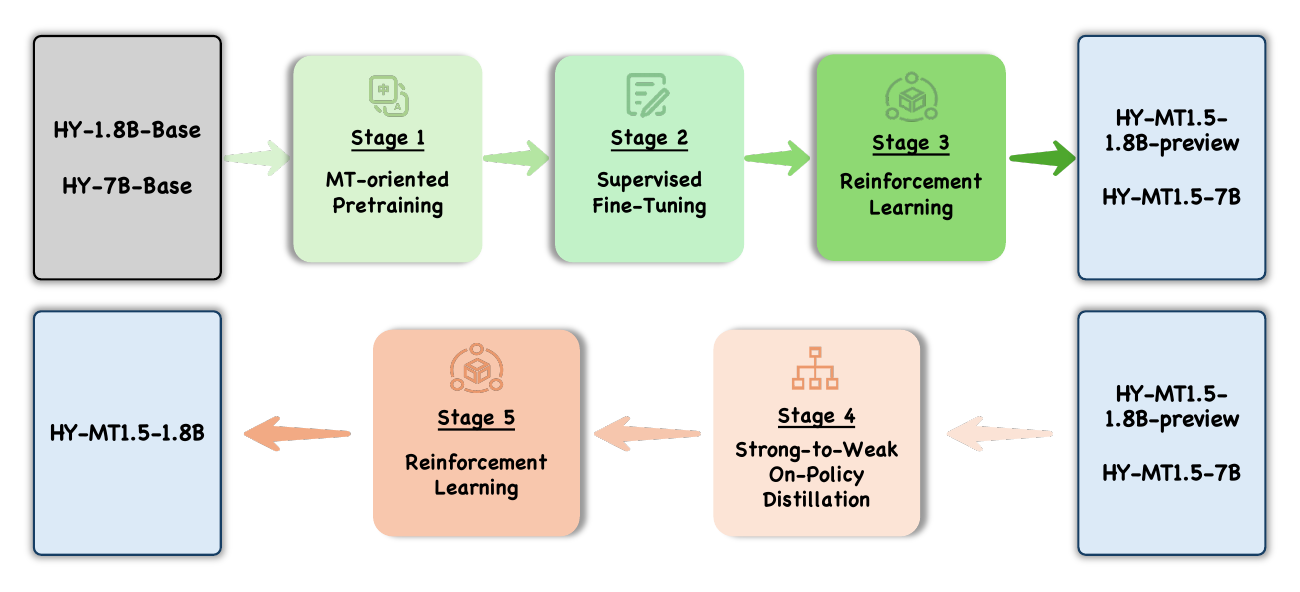}
	\caption{Training pipline of HY-MT1.5-1.8B and HY-MT1.5-7B.}
	\label{fig:pipline}
\end{figure}

Our training framework for HY-MT1.5-1.8B follows a multi-stage pipeline designed to maximize the performance of smaller-parameter models through knowledge transfer and rigorous alignment. The overall pipeline consists of four main stages: MT-oriented Pre-training, Supervised Fine-Tuning (SFT), On-Policy Distillation, and Reinforcement Learning (RL).

\subsection{MT-oriented Pretraining and Supervised Fine-Tuning}
The initial phases of our training strategy align with the methodology described in our previous Hunyuan-MT Technical Report~\citep{hunyuan-mt}. We use the HY-1.8B-Base\footnote{\url{https://huggingface.co/tencent/Hunyuan-1.8B-Pretrain}} and HY-7B-Base\footnote{\url{https://huggingface.co/tencent/Hunyuan-7B-Pretrain}} models as our base model to obtain HY-MT1.5-1.8B-preview and HY-MT1.5-7B.

\begin{itemize}
\item \noindent\textbf{Data Strategy.} We curate a massive dataset comprising high-quality multilingual monolingual corpora and parallel texts.
\item  \noindent\textbf{Process.} The base model undergoes Continuous Pretraining (CPT) followed by Supervised Fine-Tuning (SFT).
\item \noindent\textbf{Objectives.} These stages are designed to enhance the model's multilingual domain knowledge, translation capabilities, and adherence to translation instructions.
For further details on the data curation and base training protocols, please refer to our previous work \citep{hunyuan-mt}.
\end{itemize}

\subsection{Reinforcement Learning}
To further align the model with human preferences and refine translation quality, we employ Reinforcement Learning. We adopt the GRPO (Group Relative Policy Optimization) \citep{grpo} algorithm, which is also used in Hunyuan-MT-7B. GRPO updates the policy based on relative comparisons within groups of outputs, reducing training complexity while maintaining optimization stability.

We improve the reward modeling in the RL training of HY-MT1.5. Instead of relying on a single holistic score, we introduce a Rubrics-based Evaluation System. This multi-dimensional evaluation guides the LLM to evaluate translations with greater granularity.

We construct a structured scoring criterion set where an LLM-based evaluator scores translations across the following key dimensions:

\begin{itemize}
\item Accuracy: Evaluates whether the translation remains faithful to the original semantics, ensuring there are no omissions, mistranslations, or hallucinations.
\item Fluency: Assesses whether the linguistic expression is natural and conforms to the grammar and idiomatic usage of the target language.
\item Consistency: Checks for the consistent use of terminology, style, and context throughout the text.
\item Cultural Appropriateness: Examines whether the translation adapts appropriately to the cultural background and expression habits of the target language.
\item Readability: Evaluates how easy the text is to understand, ensuring clear sentence structures and distinct hierarchy.
\end{itemize}
Each dimension is assigned specific scoring standards and weights. The scores from these dimensions are aggregated to form the final reward signal. This fine-grained feedback mechanism provides the RL process with richer, more precise signals, enabling the model to improve simultaneously across multiple facets—resulting in translations that are not only correct but also natural, coherent, and culturally context-aware.

\subsection{Strong-to-Weak On-Policy Distillation}
While CPT and SFT significantly improve the 1.8B model's performance, a performance gap relative to our larger HY-MT-7B model remains due to the inherent limitations imposed by its parameter size. To bridge this gap, we employ on-policy distillation.

Recent research \citep{gkd, miniLLM, tm_on_policy} suggests that on-policy distillation is more effective than off-policy methods for improving student models. Consequently, we adopt this approach after SFT.
\begin{itemize}

\item \noindent\textbf{Teacher Model.} We utilize the fully trained HY-MT1.5-7B as the teacher model.
\item \noindent\textbf{Data.} We collect approximately 1 million monolingual samples, covering all 33 supported languages, including specific ethnic minority languages and dialects.
\item \noindent\textbf{Loss Function.} We employ per-token reverse KL divergence to align the student's output distribution with the teacher's. The loss function is defined as:
    \[
    KL(\pi_\theta \parallel \pi_{\text{teacher}}) = \mathbb{E}_{x \sim \pi_\theta} \left[ \log \pi_\theta (x_{t+1} \mid x_{1..t}) - \log \pi_{\text{teacher}} (x_{t+1} \mid x_{1..t}) \right]
    \]

\end{itemize}

This process enables the 1.8B model to inherit the 7B model's superior translation performance.
Upon completion of this phase, we employ the same reinforcement learning approach utilized in the third stage to optimize the model, yielding the final model.

\subsection{Quantization}
Recent advancements in large language models (LLMs) have demonstrated remarkable success across a wide range of applications, from conversational chatbots to creative writing. However, growing concerns over data privacy, the need for offline functionality, and the high costs of large-scale cloud deployment necessitate the direct deployment of these models on edge devices, which are typically resource-constrained. Quantization has emerged as a promising technique to achieve this goal by reducing model size and computational requirements through the use of lower-precision representations of model weights.

For HY-MT1.5-1.8B, the adoption of the W8A8C8-FP8 strategy effectively meets accuracy requirements, as FP8 provides strong support for LLM precision. For even lower bit-widths, the Weight-Int4 quantization strategy can further compress the 1.8B model to occupy less memory, catering to more demanding edge device constraints—though this comes at the cost of significant accuracy degradation. After comparing various quantization algorithms, we selected GPTQ \citep{gptq} as the post-training quantization (PTQ) calibration strategy to minimize quantization error. This algorithm processes model weights layer by layer, leveraging a small amount of calibration data to minimize the reconstruction error of quantized weights, adjusting weights via an optimization process that approximates the inverse Hessian matrix. The workflow does not require model retraining; only a small calibration dataset is needed to quantize the weights, thereby improving inference efficiency and lowering the barrier to deployment.

Extremely low-bit quantization (e.g., 2-bit, 1.58-bit) has recently attracted considerable interest from researchers and shows great potential. While ultra-low-bit quantization can compress models to an extreme degree, it also leads to substantial performance degradation. To mitigate this accuracy loss, we employ quantization-aware training (QAT) to reduce precision-related degradation through training. Unlike traditional QAT approaches, and considering the distribution characteristics of smaller models, we introduce an offset to better align the distribution of quantized weights with the original pre-quantization distribution. For 2-bit quantization, we apply symmetric quantization with a bias to achieve better results, while adopting per-channel granularity to ensure both inference performance and quantization accuracy. The weights for the model HY-MT1.5-1.8B-2BIT will be released in the near future.

\section{Experiments}



\subsection{Automatic Metrics}

To comprehensively evaluate the multilingual translation capabilities, we conducted extensive experiments using the following test sets:
\begin{itemize}
	\item \textbf{Flores-200}\footnote{\url{https://huggingface.co/datasets/Muennighoff/flores200}} \citep{flores-200}. We select 1,056 language pairs across 33 different languages from the Flores-200 dataset. These pairs are systematically categorized into three groups: Chinese $\Leftrightarrow$ XX, English $\Leftrightarrow$ XX, and XX $\Leftrightarrow$ XX translations.
	\item \textbf{WMT25}\footnote{\url{https://github.com/wmt-conference/wmt25-general-mt/blob/main/data/wmt25-genmt-humeval.jsonl}} \citep{kocmi-etal-2025-findings}. We incorporate human evaluation sets from WMT25 with 13 language pairs (Czech to German, Ukrainian and English to Bhojpuri, Czech, Egyptian Arabic, Estonian, Icelandic, Japanese,  Maasai (Kenya), Russian, Serbian (Cyrillic script), Simplified Chinese, Ukrainian). 
	\item \textbf{Mandarin$\Leftrightarrow$Minority Testset}. This test set comprises translations between Chinese and minority languages, namely Tibetan, Mongolian, Uyghur, and Kazakh.
\end{itemize}

For automatic evaluation, we use the neural metrics XCOMET-XXL \citep{xcomet} and CometKiwi \citep{cometkiwi}, which generally correlate with human judgments.

\begin{table}[ht!]
\centering
\caption{Performances of state-of-the-art models on Flores-200 \citep{flores-200}, WMT25 \citep{kocmi-etal-2025-findings}, and Mandarin$\Leftrightarrow$Minority translation. 
Specifically, we report the Chinese-centric (ZH $\Leftrightarrow$ XX), English-centric (EN $\Leftrightarrow$ XX), XX $\Leftrightarrow$ XX, and Mand. $\Leftrightarrow$ Min. performances of HY-MT1.5-1.8B, HY-MT1.5--7B, and prominent existing systems. 
Here, Mand. $\Leftrightarrow$ Min. denotes Mandarin $\Leftrightarrow$ Minority translation.
Values denoted with \(^*\) indicate that the metric scores for the corresponding model are computed only for the supported language pairs; approximately half of the total languages are unsupported.
Values replaced by \(-\) indicate that the model does not support the language pairs of the corresponding test set.
Models with open-source weights are marked with $^\dagger$. 
\textbf{HY-MT1.0-7B} refers to our previous model, Hunyuan-MT-7B\footnotemark.
Baselines are categorized into three groups: (1) {\color{blue}ultra-large general models}, (2) {\color{orange}medium to small-sized general models}, and (3) {\color{purple}translation-specialized models}.
}
\scriptsize
\renewcommand{\arraystretch}{1.2}
\setlength{\tabcolsep}{12pt}
\begin{tabular}{llccccc}
\toprule
\multirow{2}{*}{\textbf{Models}} & \multirow{2}{*}{\textbf{Metrics}} & \multicolumn{3}{c}{\textbf{FLORES-200}} & \multirow{2}{*}{\textbf{WMT25}} & \multirow{2}{*}{\textbf{Mand.$\Leftrightarrow$Min.}} \\
\cmidrule(lr){3-5}
& & \textbf{ZH $\Leftrightarrow$ XX} & \textbf{EN $\Leftrightarrow$ XX} & \textbf{XX $\Leftrightarrow$ XX} & & \\
\midrule
\color{blue}Gemini 3.0 pro & XCOMET-XXL & 0.8982 & 0.9413 & 0.8773 & 0.5505 & 0.5921 \\
\citep{gemini_3pro} & CometKiwi & 0.7882 & 0.8809 & 0.7530 & 0.6552 & 0.5274 \\
\midrule
\color{blue}DeepSeek-V3.2$^\dagger$ & XCOMET-XXL & 0.8752 & 0.9231 & 0.8421 & 0.5013 & 0.5133 \\
\citep{dsv3.2} & CometKiwi & 0.7798 & 0.8736 & 0.7521 & 0.6353 & 0.5253 \\
\midrule
\color{orange}Qwen3-235B-A22B$^\dagger$ & XCOMET-XXL & 0.8539 & 0.9029 & 0.8018 & 0.4375 & 0.4493 \\
\citep{qwen3} & CometKiwi & 0.7651 & 0.8586 & 0.7313 & 0.5820 & 0.4456 \\
\midrule
\color{orange}Qwen3-32B$^\dagger$ & XCOMET-XXL & 0.8185 & 0.8670 & 0.7433 & 0.3605 & 0.4110 \\
\citep{qwen3} & CometKiwi & 0.7429 & 0.8329 & 0.6965 & 0.5016 & 0.3841 \\
\midrule
\color{orange}Phi-4-mini-3.8B$^\dagger$ & XCOMET-XXL & 0.5839 & 0.6284 & 0.4606 & 0.2357 & 0.3542 \\
\citep{Phi-4-Mini} & CometKiwi & 0.4327 & 0.6182 & 0.3482 & 0.2819 & 0.2003 \\
\midrule
\color{purple}Tower-Plus-72B$^\dagger$ & XCOMET-XXL & 0.7969 & 0.8416 & 0.7002 & 0.4100 & 0.3855 \\
\citep{towerplus} & CometKiwi & 0.7182 & 0.8113 & 0.6553 & 0.5554 & 0.3540 \\
\midrule
\color{purple}Seed-X-PPO-7B$^\dagger$ & XCOMET-XXL & 0.7856 & 0.8312 & 0.6896 & 0.4783 & 0.4206 \\
\citep{seedx} & CometKiwi & 0.7145 & 0.8160 & 0.6436 & 0.6623 & 0.4861 \\
\midrule
\color{purple}GemmaX2-28-2B-v0.1$^\dagger$ & XCOMET-XXL & 0.7142 & 0.8208 & 0.6376 & 0.2679 & 0.3596 \\
\citep{gemmaX2-28-9B} & CometKiwi & 0.6746 & 0.8095 & 0.6310 & 0.3750 & 0.3981 \\
\midrule
\multirow{2}{*}{\color{purple}Google-Translator} & XCOMET-XXL & 0.6929 & 0.7700 & 0.6225 & 0.3893 & 0.3338 \\
 & CometKiwi & 0.6169 & 0.7552 & 0.5947 & 0.5938 & 0.3209 \\
\midrule
\multirow{2}{*}{\color{purple}Baidu-Translator} & XCOMET-XXL & 0.7690 & 0.8209 & 0.6807 & 0.2662 & -- \\
 & CometKiwi & 0.6789 & 0.7770 & 0.6369 & 0.3284 & -- \\
\midrule
\multirow{2}{*}{\color{purple}iFLYTEK-Translator} & XCOMET-XXL & 0.8196 & 0.8397 & 0.7467 & 0.2742 & 0.5011 \\
 & CometKiwi & 0.7326 & 0.8035 & 0.6868 & 0.4747 & 0.4871 \\
\midrule
\multirow{2}{*}{\color{purple}Doubao-Translator} & XCOMET-XXL & 0.8091 & 0.8673 & 0.7653 & 0.3567 & 0.5314$^*$ \\
 & CometKiwi & 0.7156 & 0.8349 & 0.6993 & 0.5869 & 0.4061$^*$ \\
\midrule
\multirow{2}{*}{\color{purple}Microsoft-Translator} & XCOMET-XXL & 0.8234 & 0.8917 & 0.8007 & 0.3993 & 0.5196 \\
 & CometKiwi & 0.7297 & 0.8546 & 0.7253 & 0.5994 & 0.3218 \\
\midrule
\multirow{2}{*}{\color{purple}Youdao-Translator (Lite)} & XCOMET-XXL & 0.8474$^*$ & 0.8930$^*$ & 0.7840$^*$ & 0.4499$^*$ & 0.4525$^*$ \\
 & CometKiwi & 0.7720$^*$ & 0.8656$^*$ & 0.7599$^*$ & 0.6520$^*$ & 0.5050$^*$ \\
\midrule
\multirow{2}{*}{\textbf{HY-MT1.0-7B}$^\dagger$} & XCOMET-XXL & 0.8643 & 0.9065 & 0.7829 & 0.6023 & 0.6082 \\
 & CometKiwi & 0.7913 & 0.8610 & 0.7210 & 0.6735 & 0.4162 \\
\midrule
\multirow{2}{*}{\textbf{HY-MT1.5-1.8B}$^\dagger$} & XCOMET-XXL & 0.8361 & 0.8942 & 0.7840 & 0.5308 & 0.5806 \\
 & CometKiwi & 0.7655 & 0.8411 & 0.7182 & 0.6195 & 0.4084 \\
\midrule
\multirow{2}{*}{\textbf{HY-MT1.5-7B}$^\dagger$} & XCOMET-XXL & 0.8690 & 0.9093 & 0.8098 & 0.6159 & 0.6174 \\
 & CometKiwi & 0.7924 & 0.8650 & 0.7336 & 0.6885 & 0.4455 \\
\bottomrule
\end{tabular}
\label{tab:translation_results}
\end{table}
\footnotetext{\url{https://github.com/Tencent-Hunyuan/Hunyuan-MT}}

As presented in Table \ref{tab:translation_results}, our experimental results indicate that the proposed HY-MT1.5-1.8B and HY-MT1.5-7B models achieve competitive performance across the XCOMET-XXL and CometKiwi evaluation metrics. 
On FLORES-200, HY-MT1.5-7B performs well across translation directions: it scores 0.8690 in ZH $\Leftrightarrow$ XX, outperforming translation-specialized models like iFLYTEK-Translator (0.8196) and Doubao-Translator (0.8091), and matching medium-sized general models such as Qwen3-235B-A22B (0.8539). 
Its 0.9093 score in EN $\Leftrightarrow$ XX surpasses most translation-specialized models and matches Qwen3-235B-A22B (0.9029), while its 0.8098 in XX $\Leftrightarrow$ XX outperforms all evaluated translation-specialized models.
On the WMT25 benchmark, HY-MT1.5-7B achieves an XCOMET-XXL score of 0.6159, significantly outperforming all compared models across all three baseline categories. 
This result is 0.0654 higher than that of the top-performing ultra-large general model, Gemini 3.0 Pro (0.5505), and far exceeds that of translation-specialized models such as Seed-X-PPO-7B (0.4783) and Tower-Plus-72B (0.4100). 
Even the smaller HY-MT1.5-1.8B model achieves an XCOMET-XXL score of 0.5308 on WMT25, outperforming many medium to small-sized general models (e.g., Qwen3-32B: 0.3605) and translation-specialized models.

A particularly noteworthy advantage is the exceptional performance of the HY-MT1.5 models on Mand. $\Leftrightarrow$ Min. translation pairs, a critical task for Chinese-centric multilingual translation. 
HY-MT1.5-7B achieves an XCOMET-XXL score of 0.6174 in this direction, which outperforms all evaluated baselines. 
It exceeds the top-performing ultra-large general model Gemini 3.0 Pro by 0.0253 (0.6174 vs. 0.5921). 
Even the 1.8B variant (0.5806) outperforms most baselines in this setting, including ultra-large models like DeepSeek-V3.2 (0.5133) and translation-specialized models such as iFLYTEK-Translator (0.5011).

The HY-MT1.5 models balance superior performance with high parameter efficiency. 
For example, HY-MT1.5-7B (7B parameters) outperforms the larger Tower-Plus-72B (72B) across FLORES-200 and Mand. $\Leftrightarrow$ Min. translation. 
Both models also outperform commercial translators (e.g., Google-Translator) and smaller general models (e.g., Qwen3-32B), validating our model design.

Moreover, HY-MT1.5-7B outperforms HY-MT1.5-1.8B across all tasks, with the largest gain (16.0\%) on WMT25, indicating that moderate model scaling boosts translation quality. 
As open-source models, they enable broader academic and industrial adoption than closed-source alternatives such as Gemini 3.0 Pro.

\subsection{Human Evaluation}

\begin{table}[ht!]
	\centering
	\renewcommand{\arraystretch}{1.1}
	\setlength{\tabcolsep}{11pt}
	\caption{Human evaluation of translation quality for the Chinese-to-English (ZH $\Rightarrow$ EN) and English-to-Chinese (EN $\Rightarrow$ ZH) directions.
    The highest scores are shown in bold.}
	\label{tab:human_evaluation}
	\begin{tabular}{lccc}
		\toprule
		\textbf{Model} & \textbf{ZH$\Rightarrow$EN} & \textbf{EN$\Rightarrow$ZH} & \textbf{Avg.} \\
		\midrule
		Baidu-Translator & 2.75 & 2.46 & 2.55 \\
		iFLYTEK-Translator & 2.88 & 2.54 & 2.65 \\
		Doubao-Translator & 2.97 & 2.48 & 2.64 \\
		Microsoft-Translator & 2.94 & 2.57 & 2.69 \\
		Google-Translator & 2.84 & 2.10 & 2.34 \\
		HY-MT1.5-1.8B & \textbf{3.01} & \textbf{2.61} & \textbf{2.74} \\
		\bottomrule
	\end{tabular}
\end{table}

To address the limitations of automatic evaluation \citep{lavie-etal-2025-findings}, we conduct human evaluation in which multilingual experts rate translations on a 0–4 scale, focusing on pre-annotated error-prone points and considering accuracy, fluency, and idiomaticity. 

As shown in Table~\ref{tab:human_evaluation}, the evaluated models are divided into two tiers: the lightweight specialized model HY-MT1.5-1.8B and mainstream commercial translation systems. 
HY-MT1.5-1.8B achieves the highest average score (2.74), outperforming all commercial systems, which is consistent with automatic evaluation. 
Most systems perform better in ZH $\Rightarrow$ EN than EN $\Rightarrow$ ZH, mainly due to the complexity of Chinese syntax generation. 

\begin{figure}[ht!]
\centering
\includegraphics[width=0.7\linewidth]{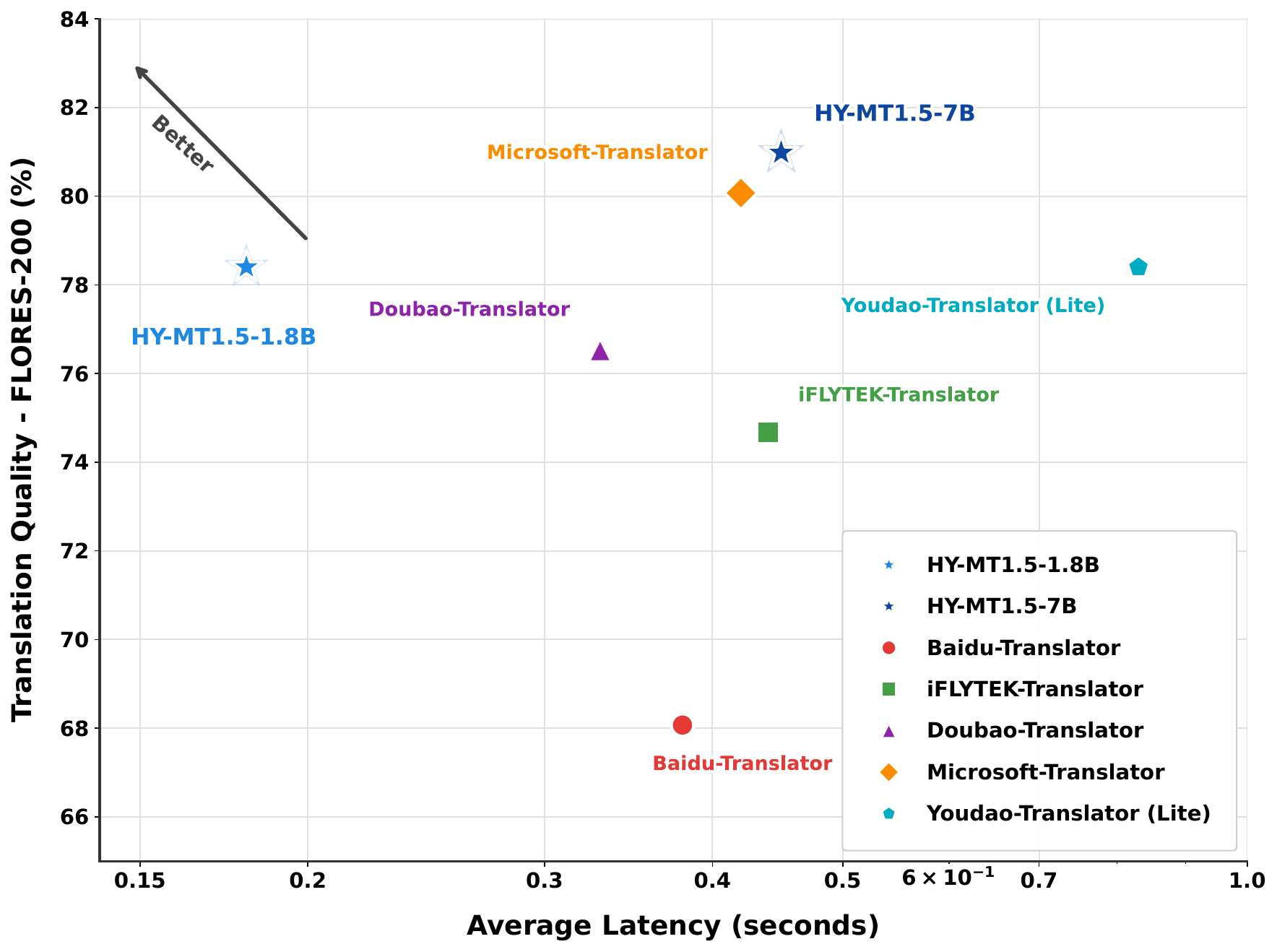}
\caption{Average response time versus translation quality for different translation models.}
\label{fig:overall}
\end{figure}
\subsection{Efficiency of HY-MT1.5 Models}

To evaluate the translation efficiency of the HY-MT1.5 models, a standardized speed test was conducted: 100 Chinese texts (average length 50 tokens, covering daily and general business scenarios) were sequentially translated into English, with the average response time used as the primary metric.

As shown in Figure \ref{fig:overall}, the HY-MT1.5 models exhibit a superior balance between translation quality and response efficiency. 
Specifically, HY-MT1.5-1.8B achieves an FLORES-200 quality score of approximately 78\% while maintaining an average response time of 0.18 seconds, indicating a clear speed advantage. 
The length of the test texts aligns with practical translation needs in daily communication, office work, and quick information retrieval, and the 0.18-second instant response time eliminates user waiting, fully meeting real-time interaction requirements.
Overall, HY-MT1.5-1.8B’s high efficiency, supported by optimized model design and inference logic, makes it well-suited for high-throughput, real-time translation scenarios such as instant messaging, intelligent customer service, and mobile translation applications.

For HY-MT1.5-7B, its quality score exceeds 80\% (exceeding that of most models in the figure), and its average response time is 0.45 seconds. This response time is comparable to that of Microsoft-Translator, whereas its translation quality is distinctly higher than that of Microsoft-Translator.
This result confirms that the HY-MT1.5 models combine high translation accuracy with fast inference speed, making them suitable for scenarios requiring both high-quality translation and real-time responsiveness.

\subsection{Practical Distinctive Features}

\begin{table}[ht!]
\centering
\renewcommand{\arraystretch}{1.4}
\setlength{\tabcolsep}{3pt}
\caption{Case studies across different scenarios, including terminology, context and format translation. }
\begin{tabular}{p{3.5cm}p{12cm}}
\hline
\multicolumn{2}{c}{\textbf{Scenario 1: Terminology Translation}} \\
\hline
\textbf{Example \#1} & 孕育出一颗混元珠 \\
\hline
\textbf{Terminology} & "混元珠": Chaos Pearl \\
\hline
\multirow{5}{*}{\textbf{Prompt Template}} & 参考下面的翻译：\\
& \{terminology\} 翻译成 \{terminology\_target\_language\}\\
& \\
& 将以下文本翻译为\{target\_language\}，注意只需要输出翻译后的结果，不要额外解释：\\
& \{source text\} \\
\hline
\textbf{Without Terminology} & Give birth to a \textcolor{red}{Hunyuan} Pearl \\
\hline
\textbf{With Terminology} & Give birth to a \textcolor{blue}{Chaos} Pearl \\
\hline
\multicolumn{2}{c}{\textbf{Scenario 2: Context Translation}} \\
\hline
\textbf{Example \#2} & The Educational and Inspirational Poetics of Pilots \\
\hline
\multirow{3}{*}{\textbf{Prompt Template}} & \{context\} \\
& 参考上面的信息，把下面的文本翻译成\{target\_language\}，注意不需要翻译上文，也不要额外解释：\\
& \{source text\} \\
\hline
\textbf{Context} & However, given the rise of the boxed DVD model that enables viewers to consume a series chronologically, viewers are now more likely to start at the beginning of a series, as they recognize that many complex television programs are designed to be watched from the start. For commercial television programs, that start is the unusual entity known as the pilot. \\
\hline
\textbf{Without Context} &  \textcolor{red}{飞行员们身上}的教育与激励性诗歌之美 \\
\hline
\textbf{With Context} & 《\textcolor{blue}{试播集}：教育与启示性的诗歌艺术》 \\
\hline
\multicolumn{2}{c}{\textbf{Scenario 3: Format Translation}} \\
\hline
\textbf{Example \#3} & \textless s1\textgreater The rain it raineth every day，\textless /s1\textgreater\textless s2\textgreater as Shakespeare noted, apparently even on Saturn. The cosmos, it seems, is no comfort at this moment.\textless /s2\textgreater\textless s3\textgreater Adam Gopnik on confronting uncertain times.\textless /s3\textgreater \\
\hline
\multirow{3}{*}{\textbf{Prompt Template}} & 将以下\textless source\textgreater \textless /source \textgreater之间的文本翻译为中文，注意只需要输出翻译后的结果，不要额外解释，原文中的\textless sn\textgreater \textless /sn\textgreater标签表示标签内文本包含格式信息，需要在译文中相应的位置尽量保留该标签。输出格式为：\textless target\textgreater str\textless /target\textgreater \\
& \\
& \textless source\textgreater\{\{\{source\_text\_with\_format\}\}\}\textless /source\textgreater \\
\hline
\textbf{Without Format} & \textcolor{red}{\textless 1\textgreater} 雨天天都在下，\textless /s1\textgreater\textless s2\textgreater 正如莎士比亚所言，显然在土星上也是如此。此刻，宇宙似乎并不能给人带来任何安慰。\textless /s2\textgreater\textless s3\textgreater 亚当·戈普尼克谈如何面对充满不确定性的时代。\textless /s3\textgreater \\
\hline
\textbf{With Format} & \textless target\textgreater\textcolor{blue}{\textless s1\textgreater} 雨日日日不停地下着，\textless /s1\textgreater\textless s2\textgreater 正如莎士比亚所言，看来在土星上也是如此。此刻的宇宙似乎并不能带来安慰。\textless /s2\textgreater\textless s3\textgreater 亚当·戈普尼克谈面对不确定时代的挑战。\textless /s3\textgreater\textless /target\textgreater \\
\hline
\end{tabular}
\label{table: scenario}
\end{table}

To address the core limitations of existing translation systems—overreliance on basic text translation and inadequate support for customized requirements via flexible prompts—we propose three distinctive features, validated through scenario-based case studies in Table \ref{table: scenario}. 
These features target terminology accuracy, context-aware disambiguation, and format preservation, which are critical for real-world scenario adaptation but underexplored in current frameworks.

First, terminology-guided translation resolves inaccurate rendering of cultural or domain-specific terms via a term-anchored prompt template. 
As shown in Scenario 1, the term “混元珠” is transliterated as the semantically ambiguous “Hunyuan” without terminology prompts; with the authorized mapping “Chaos Pearl” injected into the prompt, the model generates a precise, consistent translation.

Second, context-aware translation mitigates lexical ambiguity in contextualized tasks using a context-integrated prompt template. 
For the phrase “The Educational and Inspirational Poetics of Pilots” in Scenario 2, the model misinterprets “pilot” as “飞行员” in the absence of contextual cues; when TV series context is provided, it correctly identifies “pilot” as “试播集” and produces a contextually coherent result.

Third, format-preserving translation preserves the structural integrity of tagged text by using a format-constrained prompt template. 
In Scenario 3, the model outputs disorganized tags (e.g., \textless 1\textgreater instead of \textless s1\textgreater) without format prompts. With explicit rules to preserve \textless sn\textgreater tags and to wrap outputs in \textless target\textgreater tags, it generates translations that maintain both semantic fidelity and structural consistency.

Collectively, these prompt-driven features enable the model to transcend basic translation tasks, delivering customized solutions for terminology-sensitive, context-dependent, and format-constrained scenarios.

\subsection{Quantization Experiment}

\begin{table}[ht!]
\centering
\caption{Translation performance metrics of different quantization schemes. The highest and second-best scores are shown in \textbf{bold} and \underline{underlined}, respectively.}
{
\begin{tabular}{llccc}
\toprule
\multirow{2}{*}{\textbf{Models}} & \multirow{2}{*}{\textbf{Metrics}} & \multicolumn{3}{c}{\textbf{FLORES-200}} \\
\cmidrule(lr){3-5}
& & \textbf{ZH $\Leftrightarrow$ XX} & \textbf{EN $\Leftrightarrow$ XX} & \textbf{XX $\Leftrightarrow$ XX} \\
\midrule
\multirow{2}{*}{HY-MT1.5-1.8B} & XCOMET-XXL & \underline{0.8361} & \textbf{0.8942} & \textbf{0.7840} \\
& CometKiwi & \textbf{0.7655} & \textbf{0.8411} & \textbf{0.7182} \\
\midrule
\multirow{2}{*}{HY-MT1.5-1.8B-FP8} & XCOMET-XXL & \textbf{0.8379} & \underline{0.8905} & \underline{0.7794} \\
& CometKiwi & \underline{0.7659} & \underline{0.8396} & \underline{0.7156} \\
\midrule
\multirow{2}{*}{HY-MT1.5-1.8B-Int4} & XCOMET-XXL & 0.8060 & 0.8665 & 0.7336 \\
& CometKiwi & 0.7462 & 0.8234 & 0.6884 \\
\bottomrule
\end{tabular}
}
\label{tab:quantization_performance}
\end{table}

Large Language Models (LLMs) have achieved significant success across a range of applications, but high resource demands hinder their deployment on resource-constrained edge devices. Quantization, which reduces model size and computational cost by using low-precision weight representations, is a key solution \citep{zhang2022survey}.
For the HY-MT1.5-1.8B model, we tested two quantization strategies (Int4 and FP8) and adopted the GPTQ algorithm \citep{frantar2023gptq} for Post-Training Quantization (PTQ) to minimize errors. GPTQ processes weights layer-wise with small calibration data, avoiding retraining and improving deployment efficiency.

Table \ref{tab:quantization_performance} presents the performance of different quantization schemes (original, Int4, FP8) on multiple translation tasks, evaluated by XCOMET-XXL and CometKiwi metrics. 
It shows that FP8 quantization preserves accuracy close to the original model (e.g., ZH $\Leftrightarrow$ XX XCOMET-XXL score: 0.8379 for FP8 vs. 0.8361 for the original), whereas Int4 quantization reduces model size but causes noticeable accuracy degradation.

\section{Conclusion}

This report introduces the HY-MT1.5 models (1.8B and 7B) and a dedicated machine translation training framework that effectively addresses two core challenges: the quality-efficiency trade-off and inadequate support for customized translation needs. By integrating general pre-training, MT-oriented pre-training, supervised fine-tuning, on-policy distillation, and reinforcement learning with a rubrics-based evaluation system, the models achieve a superior balance of performance and efficiency, as well as strong adaptability to practical scenarios.

Experimental results on key benchmarks (Flores-200, WMT25, Mandarin-minority languages) confirm the competitiveness of HY-MT1.5 models. The 1.8B model outperforms mainstream medium-sized open-source models and commercial APIs, achieving performance comparable to that of ultra-large closed-source models such as Gemini-3.0-Pro, which rank in the 90th percentile. The 7B model further improves performance, surpassing Gemini-3.0-Pro on WMT25 and Mandarin-minority translation, and reaching its 95th percentile on Flores-200. With an average response time of 0.18 seconds, HY-MT1.5-1.8 B is well suited to high-throughput and real-time scenarios.

Beyond core translation performance, the models feature three practical capabilities: terminology intervention, contextual translation, and formatted translation. These prompt-enabled features flexibly meet customized demands, bridging the gap between academic research and industrial applications. Quantization experiments verify deployment potential—FP8 quantization maintains performance close to the original model while reducing resource consumption, facilitating application on resource-constrained devices.

Future work will focus on three directions: 1) Expanding language coverage to more low-resource and underrepresented languages, enhancing the inclusivity of MT technology; 2) Optimizing the training framework with more efficient knowledge distillation and reinforcement learning to improve the performance-efficiency ratio of small and medium-sized models; 3) Deepening customized translation research by integrating advanced prompt engineering and domain adaptation to better meet industry-specific needs. The HY-MT1.5 models are expected to provide high-quality, efficient, and flexible translation solutions for academic and industrial communities, advancing the development and popularization of MT technology.

\section{Contributions}
\subsection{Core Contributors}
Mao Zheng, Zheng Li, Tao Chen, Mingyang Song, Di Wang

\subsection{Contributors}
Feng Zhang, Tinghao Yu, Chengcheng Xu, Zhenyu Huang, Liya Zhan, Jun Xia, Jiaqi Zhu, Xingwu Sun, Yufei Wang, Can Xu, Liang Dong, Huxin Peng, Fei Cheng, Zheng Zhang, Liqi He, Huashuo Li, Decheng Wu, Guanghua Yu, Kai Wang, Haozhao Kuang
\end{CJK*}
\bibliography{colm2024_conference}
\bibliographystyle{colm2024_conference}
\end{document}